%% file: main.tex
\documentclass[10pt,twocolumn,letterpaper]{article}

 \usepackage{cvpr}      

\usepackage{graphicx}
\usepackage{amsmath}
\usepackage{amssymb}
\usepackage{booktabs}
\usepackage{comment}

\usepackage[pagebackref,breaklinks,colorlinks]{hyperref}
\usepackage{caption}

\renewcommand{\paragraph}[1]{\vspace{1mm}\noindent\textbf{#1}}
\usepackage{multirow}
\usepackage{tabu}

\usepackage[capitalize]{cleveref}
\crefname{section}{Sec.}{Secs.}
\Crefname{section}{Section}{Sections}
\Crefname{table}{Table}{Tables}
\crefname{table}{Tab.}{Tabs.}

\usepackage{multirow}
\usepackage[table,xcdraw]{xcolor}

\makeatletter
\@namedef{ver@everyshi.sty}{}
\makeatother
\usepackage{tikz}

\usepackage{tikz}
\usetikzlibrary{arrows,intersections}
\tikzset{font=\scriptsize}
\usepackage{pgfplots}
\pgfplotsset{compat=1.11}
\usepgfplotslibrary{fillbetween}
\usetikzlibrary{backgrounds}

\usepackage{xspace}
\usepackage{tabu}
\newlength\savewidth
\newcommand{\tablestyle}[2]{\setlength{\tabcolsep}{#1}\renewcommand{\arraystretch}{#2}\centering\footnotesize}

\usepackage{caption}
\usepackage{color, colortbl}
\usepackage{float}
\usepackage[caption=false]{subfig}
\usepackage{wrapfig}

\usepackage{url}

\definecolor{Gray}{rgb}{0.9,0.9,0.9}
\definecolor{Gray2}{rgb}{0.4,0.4,0.4}

\makeatletter
\newcommand{\thickhline}{%
    \noalign {\ifnum 0=`}\fi \hrule height 1pt
    \futurelet \reserved@a \@xhline
}
\makeatother

\usepackage{bm}
\include{math_commands}

\def\cvprPaperID{22} 
\def\confName{CVPR}
\def\confYear{2023}

\begin{document}

\title{Joint Adaptive Representations for Image-Language Learning}

\author{
AJ Piergiovanni \quad\quad\quad \quad\quad\quad \quad\quad\quad Anelia Angelova \\
Google DeepMind\\
\tt\small{\{ajpiergi, anelia\}@google.com}
}

\maketitle


\begin{abstract}
Image-language transformer models have achieved tremendous success, but they come at high computational costs. We here propose a joint adaptive image-language representation learning, which adaptively and iteratively fuses the multi-modal features.  This consistently reduces the model cost and size, allows the model to scale without a large increase in FLOPs or memory, and outperforms bigger and much more expensive models. With only 40M training examples and with 39 GFLOPs our model outperforms many times larger models, some reaching 800 GFLOPs. 


%


\end{abstract}


\section{Introduction}


%

Vision-and-language learning has made great strides recently~\cite{vilbert2020,chen2020uniter,tan2019lxmert,simvlm,git,florence,omniVL,zhang2021vinvl,li2020oscar,li2020oscar}. These models can attribute their success to scaling the well known Transformer models~\cite{attention}, which in turn need very large 
datasets. 
One important component of these models is building the underlying joint visuo-lingual representation which captures the relations between the modalities~\cite{vilbert2020,tan2019lxmert,chen2020uniter,visualbert,zhang2021vinvl,simvlm,chen2020uniter,visualbert,zhang2021vinvl,vilbert2020,tan2019lxmert,nguyen2018improved,meter,filip,albef,li2022blip,flava,clip,piergiovanni2022cotok,perceiver}.
However, expensive attention mechanisms are applied within Transformers, in which the compute required grows quadratically with the increase of the input sizes; further, these models perform better with significantly more data \cite{dosovitskiy-2021-vit} and training steps to learn the joint representations; and lastly, since large datasets are hard to collect, automatically collected datasets contain large amounts of noise.  All this makes these models even more ineffective and expensive to train: scaling the models, combined with the corresponding data scaling required, and training with large amounts of noise, require large amounts of compute. 
Thus, it is desirable to construct more memory-, FLOPs- and data- efficient vision-language representations where one can take advantage of model scale but in a more effective way. 

\begin{figure} 
\centering
   \vspace{-0.3cm}
   \includegraphics[width=1.0\linewidth]{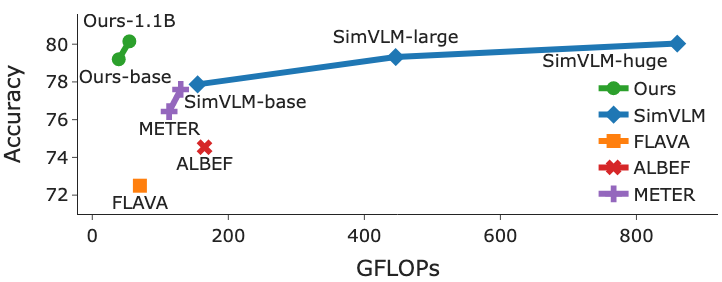}
   \vspace{-0.3cm}
    \caption{
    GFLOPs vs. accuracy for several models. The proposed approach enables much more efficient scaling, and achieves excellent performance for fewer FLOPs. It outperforms SimVLM-huge on VQA2.0 dataset, even though it is much larger and our model is evaluated in the open-vocabulary setting.
    }
      \vspace{-0.3cm}
    \label{fig:teaser}
\end{figure}

To that end, we propose the Joint Adaptive Representation for efficient image-language learning (Figure~\ref{fig:teaser}). Our approach first reduces the number of tokens in the input modalities, then adaptively fuses them. This process 
greatly reduces FLOPs, while maintaining or improving performance.
%
It results in a more compact and efficient representations, obtaining 33\% fewer FLOPs than the commonly used concatenation, while improving performance. This leads to more data- and compute- efficient models.


We evaluate the approach on Visual Question Answering (VQA) tasks, where the joint understanding of the image in the context of language input is important.
Our model performs competitively with respect to the state-of-the-art (SOTA) models, outperforming even models of large parameter and data scale (Fig.~\ref{fig:teaser}). 
Prior approaches, Perceiver~\cite{perceiver}, Co-Tokenization~\cite{piergiovanni2022cotok} also proposed efficient joint vision-language learning methods, our approach proposes a better mechanism of `updating' the information between modalities and fusing their features, surpassing these two approaches both in accuracy and in reducing FLOPs. 
%
%
Our approach allows for better model scaling, using much fewer FLOPs with increasing model sizes and input image sizes (Fig.~\ref{fig:scaling_flops}).
%
The main contribution of our work is a new image-text fusion method that is more efficient and accurate than previous methods. 
This allows us to present a novel compact image-language model of excellent performance, obtained at the fraction of the cost and data.

\section{Joint Adaptive Representations}
\label{sec:model}



\begin{figure}
    \centering
      \vspace{-0.3cm}
    \includegraphics[width=0.8\linewidth]{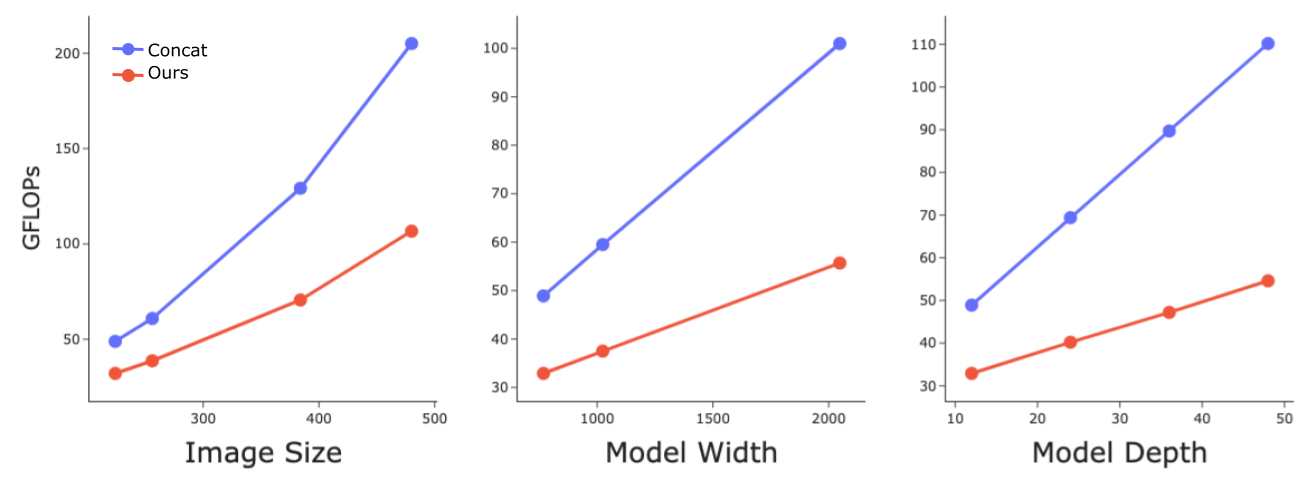}
      \vspace{-0.3cm}
    \caption{FLOPs scaling with image size, model size, and model depth (i.e., number of layers). Blue (top curve) is concatenation, red is our Joint Adaptive Representation. As seen, our approach scales more gracefully for all of them.}
      \vspace{-0.3cm}
    \label{fig:scaling_flops}
\end{figure}

The key question we address is how to combine the features from vision and language input modalities. A few basic approaches use either:
(1) concatenation or 
(2) cross-attention.
A key issue with concatenation is that it greatly increases the number of tokens by adding $H*W$ to the text length ($H$, $W$ are the height and width of the image features). Thus, as the image size increases, concatenation greatly increases the FLOPs and memory requirements of the model (Fig.~\ref{fig:scaling_flops}), e.g., ~\cite{simvlm,meter,vilbert2020,vlbert,visualbert,lu202112in1,chen2020uniter}. Here, we propose a method to reduce the number of tokens, improving efficiency.
Cross-attention based methods have other issues, mainly that the modality used for the query (usually text, e.g., ALBEF~\cite{albef}, BLIP \cite{li2022blip}),  
determines the size of the output representation. Often for vision-language tasks, the visual features have many tokens (for example, the visual tokens are 14x14 = 196 for a modest image input size of 224x224), while text is fairly short, e.g., 10 tokens in VQA2.0. When using cross-attention, the entire visual input must be squeezed into these few text token representations, greatly constraining the amount of visual information that can be used. While this approach has fewer FLOPs than concatenation, it loses information, which can reduce task performance, and puts a dependence on the input text length. Naturally, this cross-modal representation will have even less utility when increasing the input image size.

\begin{figure} 
\centering
  \vspace{-0.3cm}
  \includegraphics[width=1.05\linewidth]{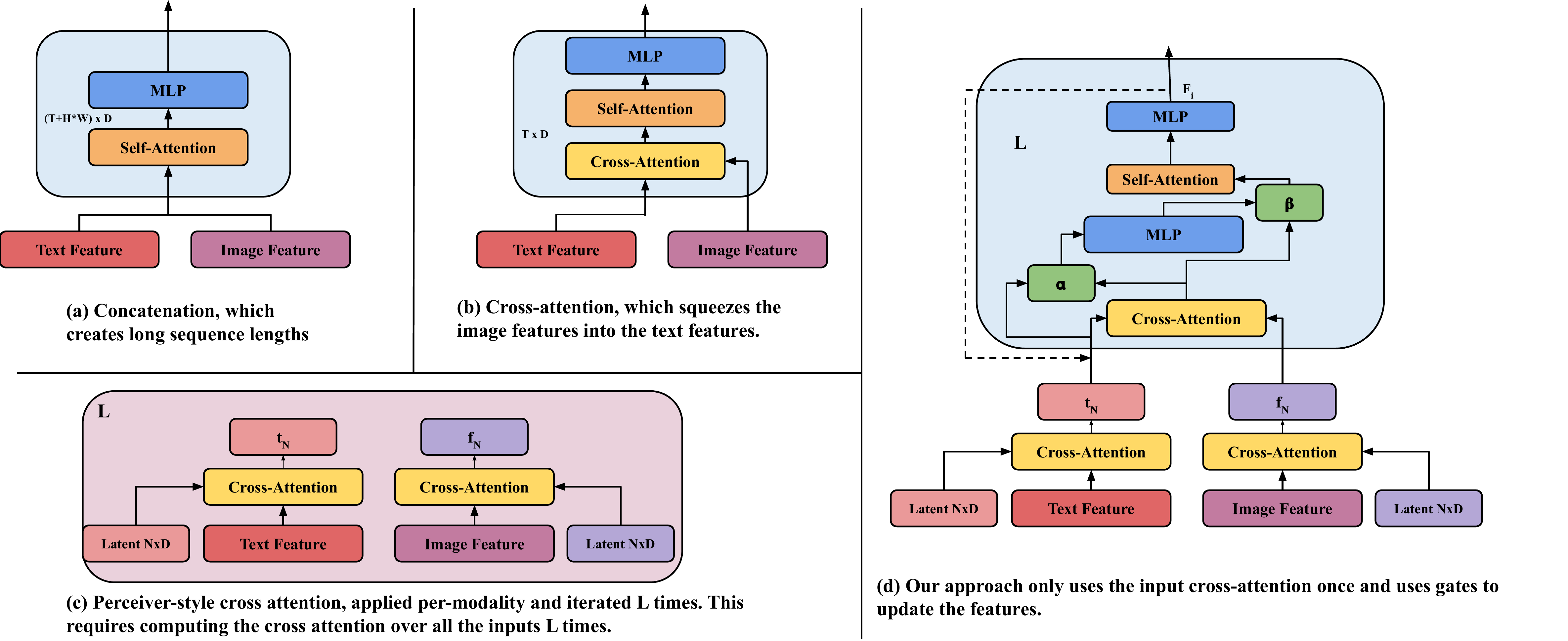}
    \vspace{-0.5cm}
    \caption{
    Visualization of the Joint Adaptive Representation (d) in the context of other approaches.
        }
          \vspace{-0.3cm}
    \label{fig:bigger_teaser}
    \vspace{-2mm}
\end{figure}

Instead, we here propose a module that enables better learning of vision-language features by more effectively incorporating the visual information and fusing it with the text information. By adaptively and iteratively tokenizing the inputs, the model is able to refine the feature representation learned from both modalities in the training process, 
while keeping a reasonable number of FLOPs (Fig.~\ref{fig:bigger_teaser}).
%

\label{sec:joint}
Our approach is based on several insights. 
First, we query the image to obtain more informative visual tokens. Previously, this was done using a TokenLearner-like approach~\cite{piergiovanni2022cotok,ryoo2021tokenlearner_neurips}. However, this method, while reducing FLOPs, notably for video applications in~\cite{piergiovanni2022cotok}, still uses quite a few FLOPs to generate and apply the attention maps, and does not scale well with image size. Instead, we utilize a hybrid approach inspired by Perceiver \cite{perceiver}. We generate $N$ tokens independently from each modality as a first step.
Secondly, we then use a direct cross-attention mechanism between the new text and compact visual features to produce a better cross-modal representation. This mechanism consists of a cross-attention layer, then a self-attention layer, and a Multi-Layer Perceptron (MLP), similar to a standard Transformer layer~\cite{attention}, but due to the reduced tokens, is much more lightweight.

Finally, this process is done iteratively, thus refining the current representation based on the set of features from the Transformer. This allows the model to dynamically update and select different visual and text features at each step so it is best able to perform the task, without increasing the compute cost. Our approach is described in detail below.

Let $X_{text}$ and $X_{im}$ be the inputs for text and for images, respectively. More specifically
$X_{text} \in \mathbb{R}^{L\times D}$ and
$X_{im} \in \mathbb{R}^{H\times W\times C}$, assuming the visual input is of size $W\times H$, the text is of length $L$. 
The goal is to produce new, lower dimensional feature representations. 
This can be done by reducing the representation to a lower number of tokens, which is particularly important for the visual features as they are many more. This is done by first unifying the representation dimensions, more specifically projecting the visual features to the $H*W\times D$ space, where $D$ is the feature dimensions for the text input: 
 \begin{equation}
 \label{eq:eq1}
 P(X_{im}) = W_1 X_{im},  
 \end{equation}
where $P(X_{im}) \in \mathbb{R}^{H*W\times D}$. Here, by $W_1$ we denote the learnable operation, e.g., applying  a  fully-connected  layer,  which  projects  the  image features  into  the D-dimensional  space.   In  principle  both the visual input and the text input can be projected to a new feature dimension e.g., thus not having to be necessarily dependent on the input feature dimension, however Eq. 1 is used here for simplicity.
In principle both the visual input and the text input can be projected to a new feature dimension e.g., $D'$, thus not having to be necessarily dependent on the input feature dimension.

As a second step, we proceed to learn a set of new $N$ learnable tokens $X_N\in R^{N\times D}$, which is done in a DETR-style~\cite{detr} feature learning. That is, $X_N$ is a randomly initialized representation that is learned via back-propagation jointly with the other parameters to minimize the loss.
\begin{equation}
\label{eq:eq1.5}
\begin{split}
f_N = &W_2 \Phi(X_{N}, P(X_{im})).   
\end{split}
\end{equation}
Here $P(X_{im})$ represents the projection of visual features from Eq.~\ref{eq:eq1}, $X_N$ is the learned latent features, $\Phi$ is the standard multi-head attention operation. This results in $f_N$,  the compact intermediate representations with $N$ features. This can also be viewed as learning $N$ new tokens, which represent the input of $M$ tokens, where $N\ll M$, for the large visual input $M=H * W$. We note that this is similar to the Perceiver architecture~\cite{perceiver}, albeit it is done only once here. This process is also done to $X_{text}$, resulting in $N$ text features ($t_N$). Thus, unlike prior work (e.g., \cite{albef,li2022blip}), $N$ is not required to be tied to the input text length; so a richer, but more compact representation is built.

Next, for the two inputs $t_N$, $f_N$ we learn a new joint feature representation $F(t_N, f_N)$ via cross attention. Importantly, we note that both these inputs will influence the subsequent representation to create a cross-modal version of text and image features.  
In the co-tokenization approach~\cite{piergiovanni2022cotok}, the two modalities are also fused for better learning, but here with two key differences: 1) the initial token reduction is not done at each iteration, which is computationally intensive; and 2) ours uses a lightweight cross-attention compared to the co-tokenization approach.


This process uses the following components.  We first use LayerNorm \cite{layernorm} (denoted as $Ln$) in order to normalize the features. We then compute cross-attention between $t_N$ (text features) and  $f_N$ (image features). The idea is that they will help construct a representation which is a combination of these modalities. We then use a standard Transformer layer with self-attention and MLPs to compute the features.
\begin{equation}
\label{eq:eq2}
\begin{split}
&P_{cr} (t_N, f_N) =  Ln(t_N) + tanh(\alpha) \Phi(Ln(t_N), Ln(f_N)) \\
&F(t_N, f_N) =  P_{cr} (t_N, f_N) + tanh(\beta) MLP(P_{cr} (t_N, f_N)) \\
\end{split}
\end{equation}
\noindent where $\alpha$ and $\beta$ are learnable parameters that control how the text and vision features are fused ($\Phi$ is the standard multi-head attention operation). We note that here, throughout, $P_{cross} (t_N, f_N) \in R^{N \times D}$, i.e., is a compact representation which combines the two modalities. We also add the tanh gating mechanism, which we find to be advantageous in our ablation experiments. 
%
The resultant representation $F(t_N, f_N) \in R^{N \times D}$ is then fed to a transformer to produce a transformed intermediate representation of the same dimension $F=\mathcal{T}(F(t_N, f_N)) \in R^{N \times D}$. We use a standard transformer layer ($\mathcal{T}$) with multi-headed attention~\cite{attention}.


\begin{table} 
     \centering
     \small
    \begin{tabular}{l|cc|ccc}
         & GFs & Data & GQA & SNLI & VQA2 \\
    \midrule
    \multicolumn{5}{c}{Large-data Models}\\
    \midrule
    
       \textcolor{Gray2}{Flamingo~\cite{flamingo}} &   \textcolor{Gray2}{ -} &  \textcolor{Gray2}{2.3B+} & \textcolor{Gray2}{-} &  \textcolor{Gray2}{-} & \textcolor{Gray2}{\textbf{82.0}} \\
       \textcolor{Gray2}{
         SimVLM~\cite{simvlm}} &   \textcolor{Gray2}{ 890$^*$} &  \textcolor{Gray2}{1.8B} & \textcolor{Gray2}{-} &  \textcolor{Gray2}{\textbf{86.21}} & \textcolor{Gray2}{80.03} \\
        \midrule
          GIT~\cite{wang2022git}  &- & 800M  &- &- & 78.81  \\
         METER \cite{meter} & 130$^*$ & 404M &  - & 80.86  & 77.68 \\
          BLIP-L \cite{li2022blip} & 250$^*$ & 129M &- &- & 78.25 \\
        \midrule
    \multicolumn{5}{c}{Small-data Models}\\
    \midrule
         FLAVA~\cite{flava} & 70$^*$ & 70M &- &78.9 &72.5 \\
         CFR~\cite{Nguyen2022Coarse} &- & - &73.6   &- &69.8 \\
         VinVL~\cite{zhang2021vinvl}    &-  &16M   &65.05 & - &75.95 \\ 
         BLIP \cite{li2022blip} & 122$^*$  & 14M &- &- & 77.54 \\
         ALBEF  \cite{albef} & 165$^*$  & 14M & -  & 80.14 & 74.54 \\
         12-in-1 \cite{lu202112in1} &- & - & 60.5 &- & 71.3 \\
         UNITER~\cite{chen2020uniter} &- & 10M &- &79.39 &72.5 \\
         LXMERT~\cite{tan2019lxmert} &- & 6.5M        &60.0 & - &69.9  \\ 

         \rowcolor{Gray}
        Ours-Base & 38.9 & 40M & \textbf{81.9} & \textbf{82.1} & \textbf{79.20}  \\

        \rowcolor{Gray}
        Ours & 54.5 & 40M & \textbf{83.1} & \textbf{84.2} & \textbf{80.15} \\
    \bottomrule
    
    \end{tabular}
  \caption{
    We outperform or perform competitively to 
    the state-of-the-art models, despite using very few GFLOPs(GFs) and small amounts of data.
    In fact with 40M training examples and with 39 GFLOPs our small model (350M params) outperforms all methods that have used $\sim$Billion examples for pre-training. 
    Models such as ALBEF and BLIP use smaller data but use have many more FLOPs. 
    Open-vocabulary evaluation.$^*$Our calculation of FLOPs. 
    }
    \label{tab:sota}
 
\end{table}

\begin{table}[]
\centering

	\begin{tabular}{l|c|cc}
\small
         & GFLOPs &  GQA & SNLI-VE  \\
    \midrule
        Perceiver~\cite{perceiver} & 40.3 & 78.2 & 77.4  \\
        CoTokenization~\cite{piergiovanni2022cotok} & 43.8 & 78.5 & 77.5  \\
        	\rowcolor{Gray}
        Ours & \textbf{38.9} &  \textbf{79.1} & \textbf{77.9}  \\
    \bottomrule
    \end{tabular}
	\caption{Comparison to the Perceiver~\cite{perceiver} method and to the Iterative Co-tokenization~\cite{piergiovanni2022cotok} approach for image+text fusion. Both are our implementations. Base Model.
	}
	
		\label{tab:comp}
\end{table}

\begin{table}[]
\centering
		\begin{tabular}{l|c|cc}
& GF & GQA & SNLI-VE  \\
		\midrule
		Concat (Baseline) & 58.4 & 78.9 & 77.4 \\
		Ours (no Gating)  &  \textbf{38.9} & 78.5 & 77.2 \\
		\rowcolor{Gray}
		Ours  & \textbf{38.9} & \textbf{79.1} & \textbf{77.9}  \\
		\bottomrule
		\end{tabular}
		\caption{Comparison to the concatenation baseline: our approach is more accurate and reduces FLOPs 1.5x. This has larger implications as most vision-language models are concatenation based.  
		}
		\label{tab:ablation_xattn} 
		\end{table}

\begin{table*} [t] 

		\captionsetup[subfloat]{width=.17\linewidth,captionskip=0pt}
	    \subfloat[\textbf{Number of Iterations} used to compute tokens.
		\label{tab:ablation:iterations}]{\tablestyle{2pt}{1.05}
		\begin{tabular}{l|c|cc}
		& GF & GQA & SNLI-VE  \\
		\midrule
		1 & 34.2 & 78.3 & 77.1 \\
		2 & 35.5 & 78.8 & 77.6 \\
		\rowcolor{Gray}
		4 & 38.9 & 79.1 & 77.9\\
		8 & 42.5 & 79.2 & 77.6 \\
		\bottomrule
		\end{tabular}}  \hspace{0.5cm}
		\subfloat[\textbf{Number of Tokens} used in the model. 
		\label{tab:ablation:tokens}]{\tablestyle{2pt}{1.05} 
		\begin{tabular}{l|c|cc}
		& GF & GQA & SNLI  \\
		\midrule
		16 & 18.5  & 76.5 & 75.8 \\
		32 & 28.4  & 78.3 & 76.8 \\
		\rowcolor{Gray}
		64 & 38.9 & 79.1 & 77.9 \\
		128 & 72.9 & 79.2 & 78.1 \\
		\bottomrule
		\end{tabular}} \hspace{0.5cm}
		\subfloat[\textbf{Resampling Method} Latent cross-attention is better. 
		\label{tab:ablation:fusion}]{\tablestyle{2pt}{1.05}
		\begin{tabular}{l|c|cc}
		& GF & GQA & SNLI  \\
		\midrule
		Spatial & 42.5  & 78.9 & 77.4 \\
		\rowcolor{Gray}
		Latent & 38.9 & 79.1 & 77.9\\
		\bottomrule
		\end{tabular}} \hspace{0.5cm}
			\subfloat[\textbf{Iterative Combination} of features after each iteration.
		\label{tab:ablation:iterative-combine}]{\tablestyle{2pt}{1.05}
		\begin{tabular}{l|c|cc}
		& GF & GQA & SNLI  \\
		\midrule
		None & 38.9 & 78.1 & 76.5 \\
		Residual & 38.9 & 78.7 & 77.6 \\
		\rowcolor{Gray}
		Weighted & 38.9 & 79.1 & 77.9 \\
		\bottomrule
		\end{tabular}} \hspace{0.5cm}
			\subfloat[\textbf{Number Layers} used  in the fusion module.
		\label{tab:ablation:layers}]{\tablestyle{2pt}{1.05}
		\begin{tabular}{l|c|cc}
		& GF & GQA & SNLI  \\
		\midrule
		8 & 22.4 & 76.7 & 74.2 \\
		16 & 30.5 & 78.3 & 75.4 \\
		\rowcolor{Gray}
		32 & 38.9 & 79.1 & 77.9\\
		\bottomrule
		\end{tabular}}
	\caption{Ablation studies exploring variants of our proposed approach. 
	}
		\label{tab:ablations}
		\vspace{-4mm}
\end{table*}

This new feature representation can be further refined to produce even better cross-modal learning by repeating the same process, but this time taking the already obtained feature as input. The operation is the same as Eq.~\ref{eq:eq2} but with continually updated input by replacing $t_N$ with $F+t_N$, which adds in the output of the previous Transformer layer. This lets the model continually refine and fuse the features. Assuming $F_i$ is the current representation and $F_{i+1}$ is the next, this uses the previous equations to iteratively update the features as follows:
\begin{equation}
\label{eq:eq4}
\begin{split}
&P_{cr} (F_i+t_N, f_N) =  Ln(F_i+t_N)+ \\ 
&\hspace{2cm} tanh(\alpha) \Phi(Ln(F_i+t_N), Ln(f_N)) \\
&F_{i+1} =  P_{cr} (F_i+t_N, f_N) + \\ 
&\hspace{2cm} tanh(\beta) MLP(P_{cross} (F_i+t_N, f_N) \\
&F_{i+1} =  \mathcal{T}(F_{i+1}) 
\end{split}
\end{equation}
Text is used at the first iteration, joint features afterwards.

Of key importance is that during the cross-modal learning process, we use the interaction of both modalities. Specifically, we use attention to determine lower dimensional projections from both modalities which differs both from the Transformer~\cite{attention} which preserves the input dimensionalities, and is a more efficient process than the Iterative Co-Tokenization~\cite{piergiovanni2022cotok} and Perceiver \cite{perceiver}, also used by Flamingo~\cite{flamingo}, as the expensive tokenization step over the whole input is only done once here. Further, different from Flamingo are the iterative updates, Eq. \ref{eq:eq4}, where we iteratively combine the features, rather than relying only on cross-attention. 
The approach is also different from methods like TokenLearner which is only applied on a single input, which can lead to a loss in accuracy if not placed appropriately \cite{ryoo2021tokenlearner_neurips}. It is also different from cross-attention methods \cite{albef,li2022blip,meter} due to the initial feature learning and iterative updating of the cross-modal information (Eq. \ref{eq:eq1.5}). 
%
This approach also offers better performance than the concatenation baselines 
while using at least \textbf{33\% fewer}  FLOPs. 

\textbf{Pre-training.}
We find that a mixture of a number of cross-modal tasks~\cite{albef,meter,piergiovanni-2022-OApretr} is more beneficial for pre-training of our vision-language model. Inspired by curriculum learning, we adaptively change the mixture ratios between the tasks during pre-training (please see the supp material for a full list of tasks and detailed description). 


\section{Experiments}

We evaluate our approach on three VQA datasets \textbf{VQA2.0} \cite{agrawal2015vqa}, \textbf{GQA} \cite{hudson2019gqa}, and Visual Entailment (\textbf{SNLI-VE} \cite{snli-ve}) where we follow the standard accuracy metrics. Our model uses the open-ended generated text which a more challenging scenario to many previous works who used fixed (3K) vocabulary and a classification setting.
Table~\ref{tab:sota} shows the comparison with the state-of-the-art (SOTA) approaches. 
We see that our method performs competitively or outperforms prior models. Of note is that both our base and our larger model are actually the lowest FLOPs among contemporary models and outperforming models with many more FLOPs (Our models use \textbf{2-20x fewer FLOPs}). 
Our small model (300M params) outperforms all SOTA approaches with the exception of extremely large models, Flamingo, SimVLM, both of which pre-train on very large datasets. Our main model further outperforms SimVLM on VQA2.0. 
Comparing to contemporary methods in terms of GFLOPs, our approach takes 38.9-54.5 GFLOPs, which is much smaller than others, e.g. ALBEF~\cite{albef} of about 165, or BLIP ranging from 120 to 250, and much smaller than SimVLM  which is close to 900 GFLOPs. 
While FLOPs is an imperfect measure, it is preferred due to differences in implementations and hardware used by other methods. Our method \textbf{reduces memory by 40\%},  memory was reduced from 15GB of the concat baseline to 9GB for ours.
%

\begin{wrapfigure}{r}{0.25\textwidth}
    \vspace{-0.65cm}
\centering
\hspace{-0.5cm}  \includegraphics[width=0.7\linewidth]{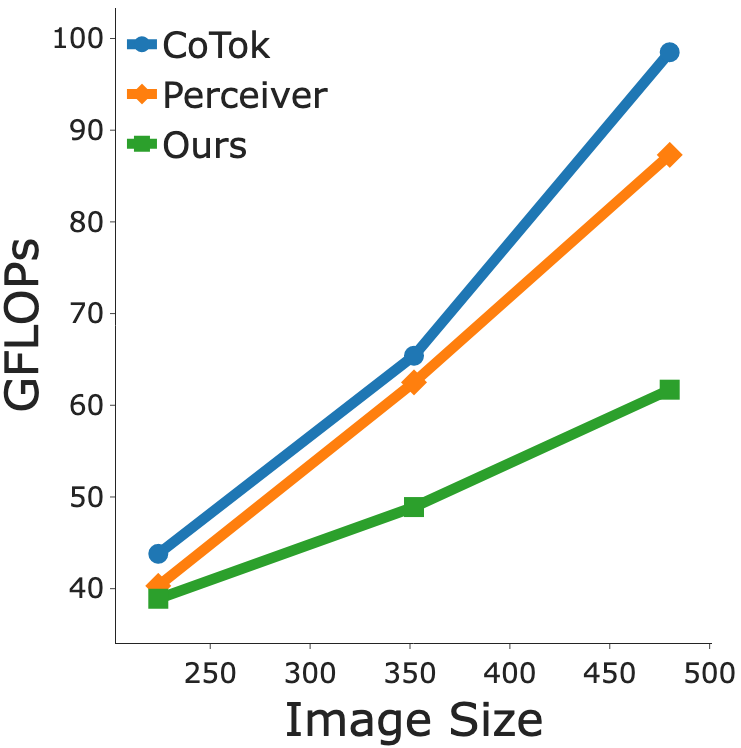}
 \vspace{-0.15cm}
  \caption{Scaling with different input sizes. With weighted iterative updates, ours scales better.}%
  \vspace{-0.25cm}
  \label{fig:sc}
\end{wrapfigure}

\textbf{Joint image-language learning comparison.} In Table~\ref{tab:comp}, we compare side-by-side our approach to other efficient image-language representation learning methods: Perceiver~\cite{perceiver} and  Iterative Co-Tokenization~\cite{piergiovanni2022cotok}. 
Our approach outperforms these advanced fusion methods, while using fewer FLOPs (Table~\ref{tab:comp}). It also scales much better than them with an increase of the input image size (Fig~\ref{fig:sc}).

\textbf{Ablation studies}
In Table \ref{tab:ablation_xattn}, we compare to the concatenation baseline, which is most commonly used ~\cite{simvlm,meter,vilbert2020,vlbert,visualbert,lu202112in1,chen2020uniter}. Our approach improves performance and reduces compute by 33\% reduction i.e. using 1.5x fewer FLOPs. 
%
Fig.~\ref{fig:scaling_flops} further shows that our approach is much more advantageous for increasing the input sizes, or model scaling, and scales better than compared to the concatenation approaches. 
%
We conduct detailed ablations to study the proposed approach. 
For each experiment, we modify one component of our main approach to verify its independent impact (`gray' is the main approach).
%
Table \ref{tab:ablations} (a) and (b) provide an ablation on iteration steps and the number of tokens learned, showing a trade-off of more compute vs higher accuracy, but with diminishing returns. 
Table \ref{tab:ablations} (c) illustrates that a single, latent cross-attention resampling of our approach gives both better performance and uses fewer FLOPs. This is in contrast to a spatial resampling used in prior works \cite{ryoo2021tokenlearner_neurips,piergiovanni2022cotok}. 
Table \ref{tab:ablations} (d), (e) study the proposed weighting (Eq. \ref{eq:eq4}), and number of layers. 
\pagebreak
{\small
\bibliographystyle{ieee_fullname}
\bibliography{egbib}
}



\end{document}

%% file: math_commands.tex
\renewcommand{\Sigma}{\mathfrak{S}}

\def\1{\bm{1}}

\DeclareMathAlphabet{\mathsfit}{\encodingdefault}{\sfdefault}{m}{sl}
\SetMathAlphabet{\mathsfit}{bold}{\encodingdefault}{\sfdefault}{bx}{n}

